\pdfoutput=1

\documentclass[11pt]{article}

\usepackage{ACL2023}

\usepackage{times}
\usepackage{latexsym}

\usepackage[T1]{fontenc}

\usepackage[utf8]{inputenc}

\usepackage{microtype}

\usepackage{inconsolata}

\usepackage{graphicx}

\usepackage{fancyhdr}

\title{Efficient In-Domain Question Answering for Resource-Constrained Environments}

\author{Isaac Chung* \and  Phat Vo* \and  Arman Kizilkale \and  Aaron Reite \\
    \\Clarifai Inc.\\
    \texttt{first.last@clarifai.com} \\
}

\begin{document}
\maketitle
\begingroup\def\thefootnote{*}\footnotetext{Equal contributions.}\endgroup
\begin{abstract}
Retrieval Augmented Generation (RAG) is a common method for integrating external knowledge into pretrained Large Language Models (LLMs) to enhance accuracy and relevancy in question answering (QA) tasks. However, prompt engineering and resource efficiency remain significant bottlenecks in developing optimal and robust RAG solutions for real-world QA applications. Recent studies have shown success in using fine tuning to address these problems; in particular, Retrieval Augmented Fine Tuning (RAFT) applied to smaller 7B models has demonstrated superior performance compared to RAG setups with much larger models such as GPT-3.5. The combination of RAFT with parameter-efficient fine tuning (PEFT) techniques, such as Low-Rank Adaptation (LoRA), promises an even more efficient solution, yet remains an unexplored area. In this work, we combine RAFT with LoRA to reduce fine tuning and storage requirements and gain faster inference times while maintaining comparable RAG performance. This results in a more compute-efficient RAFT, or \textbf{CRAFT}, which is particularly useful for knowledge-intensive QA tasks in resource-constrained environments where internet access may be restricted and hardware resources limited.

\end{abstract}

\section{Introduction}
In this paper, we propose \textbf{CRAFT}, a resource efficient approach to enhance in-domain Question Answering (QA) tasks in resource-constrained environments by combining Retrieval Augmented Fine Tuning (RAFT) \cite{zhang2024raftadaptinglanguagemodel} and Parameter-Efficient Fine Tuning (PEFT) methods, specifically Low-Rank Adaptation (LoRA) \cite{hu2021loralowrankadaptationlarge} . 
Resource-constrained environments, where systems may be isolated from unsecured networks (or "air-gapped") for security or privacy reasons are prevalent in sectors such as government, healthcare, and finance \cite{airgap1}. These settings pose significant challenges for deploying machine learning models due to restricted internet access and limited computational resources. Addressing these challenges is crucial to enable the practical application of advanced QA systems in such critical domains.

In-domain QA tasks are useful for extracting relevant information from vast datasets specific to a particular field. However, the deployment of QA models in resource-constrained environments faces significant limitations. Existing QA models typically demand high computational power for both training and inference and rely on externally-hosted models requiring continuous internet access \cite{brown2020gpt3, openai2024gpt4technicalreport, Gao2023RetrievalAugmentedGF}. These dependencies hinder the deployment of performant QA models in resource-constrained settings, where hardware capabilities are often limited and connectivity is non-existent.

To overcome these challenges, we explore the integration of RAFT and LoRA. RAFT is a technique that combines fine tuning with information retrieval, allowing the LLM to more effectively answer questions using relevant content from the retrieved data. In resource-constrained settings, RAFT compensates for reduced model capacity by expanding the model's contextual knowledge while simultaneously reducing the model's vulnerability to incorrect retrievals.

On the other hand, LoRA \cite{hu2021loralowrankadaptationlarge} is a PEFT technique designed to fine tune models efficiently by training lightweight adapters that are added into the frozen model. LoRA adapters consist of a much smaller number of trainable parameters compared to the full model, ensuring efficient storage and fine tuning while maintaining or even improving performance on QA tasks compared to fine tuning the full model (hereafter referred to as supervised fine tuning, or SFT). Furthermore, different or custom LoRA adapters can be dynamically replaced in a LLM residing in an inference server's memory immediately prior to inference, a process referred to as adapter swapping. Both RAFT and LoRA are discussed in further detail in Section 2. The individual benefits of RAFT and LoRA suggest that their combination could provide a synergistic advantage, leading to RAG systems that are both effective, resource-efficient, and highly customizable via adapter swapping.

 We focus our study on LLMs within the 7-8 billion parameter range, like Llama3 and Llama3.1 \cite{dubey2024llama3herdmodels}. This specific range is chosen to balance the trade-offs between model size, performance, and resource requirements. Larger models like GPT-4 \cite{openai2024gpt4technicalreport} generally offer better performance but at the cost of increased computational demands, which are impractical in resource-constrained environments, if not impossible, to access due to firewalls or connectivity. Conversely, smaller models may not provide the necessary performance for complex QA tasks in regular RAG systems. The 7-8B parameter range strikes an optimal balance for RAFT systems, offering sufficient capability while remaining feasible for deployment in resource-constrained settings.

The primary objectives of this research are to reduce training requirements, achieve faster inference times, enable adapter swapping, and maintain or improve the performance of LLMs in RAG systems. By combining RAFT and LoRA methods, we aim to create a framework that addresses the specific needs of resource-constrained environments without compromising on the quality of the QA system. 

To assess the effectiveness of our proposed models, we evaluate them on both their QA performance and resource utilization. The choice of these metrics ensures a comprehensive evaluation of both the efficiency and reliability of the models.

We release all models and generated datasets to facilitate further study on HuggingFace \footnote{https://huggingface.co/collections/phatvo/raft-66c44a18ac74db25de87a4dc}. In Section 2, we present some related work on RAG, fine tuning methods for RAG including RAFT, and PEFT techniques such as LoRA. In Section 3, we introduce our compute-efficient RAFT method, CRAFT, in detail. In Section 4, we present the experiment setup. In Section 5 we report the results, and in Section 6 we conclude the paper. 
\section{Related Work}

\paragraph{Retrieval Augmented Generation (RAG)}
RAG \cite{lewis2021retrievalaugmentedgenerationknowledgeintensivenlp} enhances LLMs by retrieving relevant document chunks from external knowledge bases through semantic similarity calculations. This method mitigates the generation of factually incorrect content by referencing external knowledge rather than relying solely on knowledge the model learned during training, thereby improving the relevancy of the generated text while reducing "hallucinations". Despite its advantages, RAG faces challenges, particularly with domain-specific or knowledge-intensive tasks, particularly when handling queries beyond the scope of its retrieved data \cite{zhang2023sirenssongaiocean}, though at a lesser extent when compared to non-retrieval-augmented LLMs. Other major challenges with RAG includes requiring a high-performing retriever model to produce representative embeddings from the document chunks and retrieval system that balances scale and accuracy. Recent advances in RAG have expanded its applications across various domains, showcasing its versatility and potential \cite{yan2024crag}. RAG excels in dynamic environments by offering real-time knowledge updates and effective utilization of external knowledge sources with high interpretability. However, it comes with higher latency and the possibility of added noise from extraneous contexts. 

\paragraph{Fine tuning for RAG}
Fine tuning strategies for RAG involve further training of a pretrained LLM on a specific dataset to enhance its performance in RAG tasks over that dataset. Several studies, such as those by \cite{lin2024radit} and  \cite{xu2024retrievalmeetslongcontext} have explored different fine tuning methodologies for improving LLMs in RAG tasks. These works focus on the benefits of retrieval on long context (instruction-tuned) LLMs and extending the scope of fine tuning to the retriever. 
RAFT \cite{zhang2024raftadaptinglanguagemodel} includes a fine tuning strategy that generates training data from the QA target domain data for instruction fine tuning. The entire target domain is chunked into a smaller collection of documents by a fixed token count. For each document, a larger, highly performant LLM, usually different from the LLM to be fine tuned, is used to generate a question that can be answered using the document. A Chain-of-thought (CoT)\cite{wei2023cot} style answer is then generated by providing the same, larger LLM with the context of the document and question from the previous step. This response has a full reasoning chain, i.e. a series of intermediate reasoning steps, and clearly cited sources. Then the model is fine tuned via standard supervised techniques using the generated QA pairs as well as the original, or "golden" document for context, intermixed with irrelevant or "distractor" documents. In some training instances, the "golden" document is intentionally left out, which motivates the model to directly memorize answers from the target data during fine tuning instead of relying exclusively on the provided context. This increases performance and results in a model less susceptible to incorrect RAG retrievals. Our work follows the RAFT strategy but employs LoRA instead of SFT.

\paragraph{Parameter Efficient Finetuning (PEFT)}
PEFT is a class of methods to adapt pre-trained large models to specific tasks or domains by fine tuning a much smaller set of parameters compared to SFT \cite{han2024parameterefficientfinetuninglargemodels} and may be used to enhance QA performance with or without RAG. PEFT involves selectively fine tuning a small proportion of model parameters to adapt it to a specific task or domain, or introducing new trainable parameters to the model while keeping the rest frozen. LoRA \cite{hu2021loralowrankadaptationlarge} is a notable PEFT technique that significantly decreases computational burden while maintaining performance on a wide array of tasks comparable to SFT \cite{hu2021loralowrankadaptationlarge,zhang2024autoloraautomaticallytuningmatrix}. LoRA consists of a set of pairs of low-rank matrices of size $n \times k$ and $k \times n$, where $k << n$. The multiple of each pair is added into a single corresponding matrix of size $n \times n$ within the frozen transformer model. LoRA adapters may be added into any subset of matrices in the transformer while maintaining differentiability, allowing for efficient fine-tuning as each pair contains a total of $2(n \times k)$ trainable parameters, much fewer than the $n^2$ parameters of the frozen matrix. This approach significantly reduces memory usage and computational requirements compared to fine tuning the full model, making it suitable for adapting LLMs to a new domain in resource-constrained environments with limited resources. Moreover, multiple LoRA adapters may be trained on a single frozen LLM: for example, different LoRA adapters may be trained for different datasets, or even users, and then quickly added into a frozen LLM already in an inference servers' memory, a process referred to as adapter swapping. In contrast, fully fine tuning the entire LLM for a multitude of differing use cases requires much more compute and storage, and swapping an entire LLM into an inference server's memory prior to inference introduces unacceptable latency when compared to adapter swapping.

\section{Methods}
In this section, we detail our method to train a LoRA adapter for RAFT. We largely follow the setup described in RAFT \cite{zhang2024raftadaptinglanguagemodel} and report the major differences. We use the original recipe to generate training data from the target in-domain data; then we use LoRA to fine tune an adapter on the generated dataset instead of SFT. In the remainder, the term RAFT refers to a model fine tuned with SFT, while CRAFT refers to a model fine tuned with LoRA. 

\textbf{Fine Tuning:} At the data generation stage, we follow the RAFT recipe, but we  substitute Llama3-70B-instruct for GPT-4 \cite{openai2024gpt4technicalreport} to generate one question per context that can be answered using that context. We chose Llama3-70B-instruct as a GPT-4 replacement as other proprietary models may not be accessible in resource-constrained environments. Since using this 70B model (FP16) for inference requires around 130GB, we used vLLM \cite{kwon2023vllm} as the inference server over 4x48GB NVIDIA RTX 6000 Ada GPUs. Note that this inference server is only needed at the data generation stage and not in the fine tuning or inference stages.

We include 3 distractor documents instead of the 4 used in the original RAFT paper to further reduce the number of tokens used during training, resulting in slightly more efficiency. The golden and distractor documents, along with the question and the CoT answer, are then formatted to form one instruction. 

\textbf{LoRA:} We tune the hyperparameters of the LoRA training configurations. Specifically, we tune the rank of the adapters ($r=[4,16,64]$) and report the best result in the Results section. 

\textbf{Model Selection:} We focus on pretrained LLMs within the 7-8B parameter range. These models offer a balanced trade-off between performance and computational feasibility, making them suitable for deployment in resource-constrained environments. As air-gapped environments may not be able to access external APIs, we focus our study on Llama3.1-8B-Instruct \cite{dubey2024llama3herdmodels}, which can be deployed locally. 

\section{Experiments}

In this section, we describe our experimental setup and our results, offer a detailed analysis, and highlight the importance of PEFT in minimizing GPU memory usage during training for resource-constrained systems.

\subsection{Datasets}

For our experiments, we selected the following datasets:

\begin{itemize}
    \item HotPotQA dataset \cite{yang2018hotpotqa}, which provides a diverse set of multi-hop QA pairs across various topics from Wikipedia. This differs from the other listed datasets here where the questions require finding and reasoning over multiple supporting documents to answer as opposed to only a single document (single-hop). 
    \item NarrativeQA \cite{kocisky2018narrativeqa}, which provides QA pairs derived from stories based on books and movie scripts.
    \item NewsQA \cite{trischler2017newsqa}, which provides QA pairs based on a set of over 10,000 news articles from CNN, with answers consisting of spans of text from the corresponding articles. 
    \item PubMedQA \cite{pubmedqa}, which provides QA pairs for reasoning over biomedical research texts. It mainly focuses on answering yes/no medical and biology questions based on a given set of documents.
    \item WebGLM-QA \cite{liu2023webglm}, a long-formed and properly cited QA dataset curated via LLM in-context bootstrapping to train the WebGLM generator module. 
\end{itemize} 

\textbf{Data Preprocessing:} Each dataset is segmented into context chunks, each containing relevant information that could be used to generate question-answer pairs. All of the selected datasets already have pre-determined contexts. 

\textbf{Sample Selection:} To manage computational constraints, a sample of 100 chunks is selected for generating question-answer pairs for fine tuning. The evaluation set is also subsampled to only 1000 rows to speed up computations. This selection ensures that the training process remains feasible within the resource limitations of resource-constrained environments, allowing us to focus on optimizing the model's performance without exceeding hardware capabilities.

\subsection{Baselines}
We consider the following baselines for our experiments:

\begin{itemize}
    \item \textit{Llama3.1-8B-Instruct + Golden} model represents an idealized RAG setup. In order to eliminate the effects of retrieval errors on the benchmarks the model is always provided with the golden context.
    \item \textit{Llama3.1-8B-Instruct + RAG} model represents a realistic setup where the retriever is operational; hence golden context is not guaranteed to be in the context.
\end{itemize}

For the RAG setup, we use a naive RAG pipeline, also known as a retrieve-read framework \cite{ma2023queryrewritingretrievalaugmentedlarge} that only involves retrieval and generation, and does not include any pre-/post-processing or advanced techniques. We use the \texttt{BAAI/bge-small-en-v1.5} text embedding model \cite{bge_embedding}, one of the top models on MTEB \cite{muennighoff2023mtebmassivetextembedding} with <100M parameters, to generate embeddings. The index is generated with FAISS \cite{johnson2019billion} and deployed as the retriever in the RAG pipeline, with the \texttt{top\_k} retrieval set to 5.

\section{Results}

We find that the CRAFT (a LoRA/RAFT finetuned version of Llama3.1-8B-Instruct) is better at reading and extracting information from in-domain documents, compared to a general-purpose model with RAG.

\begin{table*}
    \centering
    \begin{tabular}{lccccc}
    \hline
        ~ & HotPotQA & NewsQA & NarrativeQA & PubMedQA & WebGLM-QA \\ \hline
        Llama3.1-8B + Golden & 41.95 & 46.93 & 59.25 & 77.00 & 18.34 \\ 
        CRAFT + Golden & \textbf{48.21} & \textbf{47.14} & \textbf{64.54} & \textbf{79.67} & \textbf{36.59} \\ \hline
        Llama3.1-8B + RAG & 19.60 & 20.80 & 35.60 & 73.00 & 22.00 \\
        CRAFT + RAG & \textbf{44.70} & \textbf{30.10} & \textbf{47.20} & \textbf{75.00} & \textbf{39.00} \\ \hline
    \end{tabular}
    \caption{F1 scores on HotPotQA, NewsQA, NarrativeQA, PubMedQA, and WebGLM-QA comparing CRAFT and the baseline models: (i) an idealized RAG (Golden) and (ii) a realistic RAG. Bold numbers denote the best score in each comparison, where higher is better.}
\label{table:main-results}
\end{table*}

\begin{table*}
    \centering
        \begin{tabular}{lcc|cc|cc}
        \hline
            & \multicolumn{2}{c}{\normalsize HotPotQA} & \multicolumn{2}{c}{\normalsize NewsQA} & \multicolumn{2}{c}{\normalsize NarrativeQA} \\
            ~ & (a) & (b) & (a) & (b) & (a) & (b) \\ \hline
            Trainable Params (M) & 8100 & 168 & 8100 & 168 & 8100 & 168 \\
            Training time (min) & 180 & 9 & 240 & 41 & 100 & 19 \\ 
            GPU Memory (GB) & 40.5 & 21.0 & 40.5 & 32.0 & 40.5 & 26.0 \\
            Perplexity & 1.20 & 1.24 & 1.075 & 1.12 & 1.65 & 1.57 \\ \hline
        \end{tabular}
    \caption{Comparison of a) RAFT and b) CRAFT using generated RAFT data from HotPotQA, NewsQA, and NarrativeQA datasets. We also report perplexity over selected datasets. Lower perplexity is better. }
\label{table:ft-v-lora-ablation}
\end{table*}

Using the above datasets and baselines, we evaluate our proposed method and demonstrate the effectiveness of CRAFT in Table \ref{table:main-results}. 
Overall, the Llama3.1-8B-instruct model performs well due to its pre-training and instruction-tuned answering style. 
We see that CRAFT consistently outperforms the baselines. Compared to the Llama3.1-8B-instruct model, CRAFT does much better in terms of extracting information, yielding an average of 13.41\% gains over the evaluated datasets. 

\paragraph{Single-Hop QA vs Multi-Hop QA}
CRAFT enhances QA scenarious requiring multi-hop reasoning substantially more than those requiring only single-hop reasoning. The performance gains over the baseline model when provided the golden context are 14.9\% for HotPotQA, 4.47\% for NewsQA, and 8.93\% for NarrativeQA. These gains are dramatically amplified in RAG scenarios, to 128\%, 44.7\% and 35.6\% for HotPotQA, NewsQA, and NarrativeQA respectively. 

\paragraph{Comparison of RAFT and CRAFT}conduct an analysis to illustrate the resource efficiencies achieved by using LoRA compared to SFT for fine tuning on the same generated dataset. As shown in Table \ref{table:ft-v-lora-ablation}, using CRAFT reduces the number of trainable parameters to just 2\% of those required by SFT. This results in a nearly 35\% decrease in GPU memory usage during training and an average of speedup of 7.5x. In terms of perplexity,  for HotPotQA and NewsQA, SFT offered marginally lower perplexity (by 0.04 and 0.045, respectively), where LoRA achieved a lower perplexity for NarrativeQA (by 0.08). Overall, the average perplexity between the two methods is nearly identical, indicating very similar model performance.

\section{Conclusions and Future Work}
In this paper, we introduced CRAFT, a method that combines RAFT and LoRA to efficiently adapt LLMs in resource-constrained environments while still retaining competitive performance in knowledge-intensive QA tasks. 

Notably, CRAFT still requires a relatively large LLM for generating questions and CoT answers. A quantized model can certainly reduce memory requirements. For example, a 4-bit quantized Llama3-70B-instruct model would only require around 35GB and would fit in a 1x48GB setting. 
A potential avenue to avoid using any large models could be an ensemble of smaller models using an mixture-of-agents approach \cite{wang2024mixtureofagentsenhanceslargelanguage}. 

Investigations on how to combine retrieval augmented fine tuning methods and methods that further reduce memory usage for fine tuning or leverage quantization, such as QLORA \cite{dettmers2023qlora}, is left for future work. 

\section*{Acknowledgements}
We would like to thank Mark Lowell for providing feedback on the paper. 

\bibliography{anthology,custom}
\bibliographystyle{acl_natbib}



\end{document}